\documentclass[10pt,letterpaper]{article}

\usepackage{cogsci}

\cogscifinalcopy 

\usepackage{pslatex}
\usepackage{apacite}
\usepackage{lipsum}
\usepackage{enumitem}
\usepackage{float} 
\usepackage{graphicx}
\usepackage{xcolor}
\usepackage{url}

\usepackage[hyphenbreaks]{breakurl}

\newcommand{\gtlogo}{\raisebox{0pt}{\includegraphics[scale=0.04]{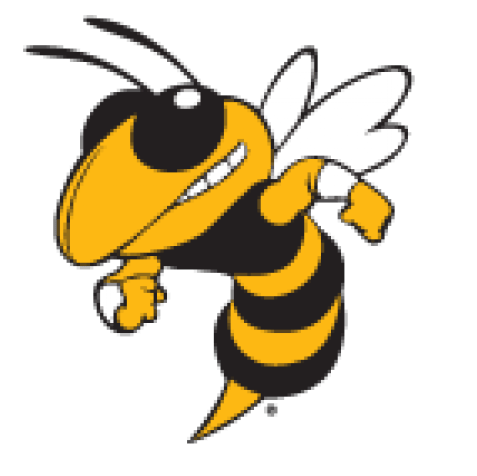}}}

\title{How Well Do Deep Learning Models Capture Human Concepts? The Case of the Typicality Effect}

 \author{{\large \bf Siddhartha K. Vemuri, Raj Sanjay Shah, Sashank Varma} \\
  \{svemuri8, rajsanjayshah,
varma\}@gatech.edu
  \\
  Georgia Institute of Technology \gtlogo
  }

\begin{document}

\maketitle

\begin{abstract}
How well do representations learned by ML models align with those of humans? Here, we consider concept representations learned by deep learning models and evaluate whether they show a fundamental behavioral signature of human concepts, the \emph{typicality effect}. This is the finding that people judge some instances (e.g., robin) of a category (e.g., Bird) to be more typical than others (e.g., penguin). 
Recent research looking for human-like typicality effects in language and vision models has focused on models of a single modality, tested only a small number of concepts, and found only modest correlations with human typicality ratings. 
The current study expands this behavioral evaluation of models by considering a broader range of language ($N = 8$) and vision ($N = 10$) model architectures. 
It also evaluates whether the \emph{combined} typicality predictions of vision + language model pairs, as well as a multimodal CLIP-based model, are better aligned with human typicality judgments than those of models of either modality alone. Finally, it evaluates the models across a broader range of concepts ($N = 27$) than prior studies. There were three important findings. First, language models better align with human typicality judgments than vision models.
Second, combined language and vision models (e.g., AlexNet + MiniLM) better predict the human typicality data than the best-performing language model (i.e., MiniLM) or vision model (i.e., ViT-Huge) alone. Third, multimodal models (i.e., CLIP ViT) show promise for explaining human typicality judgments.
These results advance the state-of-the-art in aligning the conceptual representations of ML models and humans. A methodological contribution is the creation of a new image set for testing the conceptual alignment of vision models.

\end{abstract}

\section{Introduction}

\label{intro}

Categorization is a fundamental aspect of cognition. Assigning a new stimulus to a category enables humans to make inferences about its unknown or unobservable properties, facilitating taking action in the world \cite{murphy2002big}. A classic proposal in cognitive science is that not all the members of a category have the same status \cite{rosch75}. Rather, members vary in their \emph{typicality}, with some members (e.g., robin) more typical of a category (e.g., Birds) than others (e.g., penguin). 
Moreover, people are faster to understand sentences about typical vs. atypical category members, and are quicker to give the category label (e.g., "Birds") of typical vs. atypical members when presented with images \cite{murphy2002big}.

Here, we investigate whether Large Language Models (LLMs) and computer vision (CV) models also show typicality gradients. Prior studies that have investigated this question have found suggestive results \cite{misra2021language, upadhyay}. We go beyond this work to examine a larger number of LLMs and CV models, and to evaluate these models using newer human typicality data collected over a larger number of categories. We investigate for the first time whether the \emph{combined} typicality predictions of vision + language model pairs better align with human typicality judgments than those of models of either modality alone. We also explore the potential of multi-modal models (i.e., CLIP). 
\subsection{The Typicality Effect}

The typicality effect is that people regard some members as ``better’’ examples of a category than others. Investigating typicality gradients requires collecting data from humans. The most common procedure is to give participants a category label (e.g., Fruits) and to have them write down as many exemplars of the category as they can in a fixed amount of time, usually 30 seconds \cite{battig1969category, castro21, van2004category}. The typicality of a member is defined as the proportion of participants who produce it.
Another approach is to provide a category label and a sequence of members and have participants rate the ``goodness’’ of each member on a scale ranging, for example, from 1 (very typical) to 7 (very atypical) \cite{rosch75}. The typicality of a member is its average rating across participants.

\subsection{Typicality in ML Models}

We investigate whether ML models trained on large corpora or image sets also show the typicality effect. These models learn about the statistical structure of the cognitive environment to perform word prediction tasks or image classification tasks, respectively. Here, we evaluate whether as a consequence of this training, they become sensitive to the typicality gradients that organize the members of categories, learning them as latent representations. 

\subsubsection{Typicality in Language Models}

Researchers have looked for typicality effects in language models. An early study investigated whether word2vec embeddings could be used to predict category typicality data \cite{heyman2019prediction, de2008exemplar}. The mean correlation between word2vec and humans across 16 categories was only 0.29.
A more recent study \cite{misra2021language} investigated the alignment of more modern transformer-based models (including RoBERTa and GPT-2) to the \citeA{rosch75} typicality ratings for 10 categories. The 19 models tested showed a range of correlations, with the larger variants of RoBERTa and GPT-2 achieving the highest values of approximately 0.40. \citeA{bhatia2022transformer} developed a BERT-based model and evaluated it against 25 findings on semantic cognition, including that of typicality gradients. The model’s typicality ratings across 10 categories correlated 0.32 with the human ratings of \citeA{rosch75}.
Thus, we see that NLP models have shown modest abilities to account for the typicality effect. 

\subsubsection{Typicality in Vision Models}

Researchers have also investigated whether vision models align with human conceptual understanding \cite{battleday2021from}.
An early study \cite{peterson2018evaluating} had participants rate the pairwise similarity of 120 images of exemplars from each of 6 categories. They compared these to the cosine similarities of the representations on the final fully connected layers of several CNN models, finding moderate correlations for VGG-16 \cite{simonyan2014very} in particular. Subsequent research combined the low-level visual processing of CNNs with cognitive science models of decision-making. These hybrid models were evaluated against data collected on CIFAR-10 test images \cite{battleday2020capturing, singh2020end}.

The most recent work in this area evaluated 'stacked' methods for approximating human similarity judgments \cite{marjieh2023large}
This has shown the value-added of cognitive science models but has not addressed the typicality effect. 

Most relevant is \citeA{upadhyay}, who investigated whether the CNN model VGG-19 shows typicality gradients. This proof-of-concept study focused on the well-studied Bird category, finding only small (0.32) and non-significant correlations between model-predicted and human typicality ratings. Thus, it remains an open question whether vision models can account for the typicality effect observed in humans.

\subsubsection{Research Goals}

Despite recent work, large gaps remain in our understanding of whether ML models trained on large data sets acquire, purely through experience, conceptual representations that resemble those of humans. The current study addressed these gaps through the lens of the typicality effect. There were four research goals:
\vspace{-2mm}
\begin{enumerate}[itemsep=-1.2mm]
    \item To evaluate the alignment between a large number of language models of varying architectures/sizes and recently collected human typicality ratings across a large number of categories.
    \item To do the same for a large number of vision models of varying architectures/sizes.
    \item To evaluate, for the first time, whether combining the predictions of a language model and a vision model offers a better account of human typicality than either model alone.
    \item To evaluate, for the first time, whether a multimodal model offers better predictions of human typicality than models of a single modality (vision or language).
\end{enumerate}
\vspace{-2mm}
The current study also makes two methodological contributions. The first is to evaluate ML models across a broader range ($N = 27$) of categories than has previously been considered, using human data that has been collected in the past few years rather than decades ago. The second is to develop a new set of ‘naturalistic’ images to test the conceptual alignment of vision models.

\section{Methods}
\label{methods}

\subsection{Data Preparation}

\paragraph{Human Typicality Ratings}
We used the human typicality ratings of \cite{castro21}. This is the most recent dataset of its kind, and it supersedes the norms \cite{rosch75} used in many prior studies of the alignment of language and vision models to human categorization. 250 participants provided exemplars of each of the 70 categories -- as many as possible within a 30-second time frame. The typicality of an exemplar was defined as the proportion of participants who produced it. 
We selected 27 concepts whose exemplars have concrete and distinct visual depictions, to be able to evaluate the vision models.

\paragraph{Image Collection and Processing}
Images were collected via the Google Image Search package \cite{google-images-search}. For each exemplar of each category, we used its label as a search string and collected at least 20 images (after removing corrupted images and images with unusable file formats). 
We used the CarveKit Image Background Removal Tool \cite{image-background-remove-tool} to remove the backgrounds of all images where the background was distinct from the exemplar itself, e.g., removing the sky and tree branches from an image of the robin exemplar of the Bird category. Removed backgrounds were replaced with a plain white background. Background removal was not performed for the Color, Dwelling, Earth Formation, Fabric, Tree, and Weather categories because the exemplars were often inseparable from the backgrounds. We manually reviewed the outputs of automated image collection and background removal and discarded images that did not depict the intended exemplar and those with improper background removals. These images were converted to the JPG file format, resized to 224 x 224 pixels, and normalized using the ImageNet mean and standard deviation values. Our image collection and processing code is public, but please contact the authors for access to the specific image set used for the experiments discussed in this paper. \footnote{We make all our code publicly available at \url{https://github.com/svemuri8/cv-nlp-typicality/tree/main}}

\subsection{Model Selection}

\paragraph{Language Models}

We selected several pre-trained language models spanning a range of architectures: word2vec \cite{word2vec}, GloVe \cite{glove}, RoBERTa-large \cite{roberta}, XLNet-base \cite{xlnet}, MiniLM \cite{minilm}, MPNet \cite{mpnet}, T5-large \cite{t5}, and GPT (text-embedding-ada-002) \cite{gpt}. All models lacked classification heads or decoders. Instead, they produced word/sentence embedding vectors. The GloVe and word2vec implementations were sourced from Gensim \cite{gensim} and the GPT embeddings from OpenAI's API \cite{openai}. All other models are from HuggingFace's
Transformers Library \cite{sentence-transformers}. 


\paragraph{Vision Models}

We selected several vision models pre-trained on ImageNet1K \cite{deng2009imagenet} spanning different architectures: AlexNet \cite{alexnet}, VGG19 \cite{vgg19}, InceptionV3 \cite{inceptionv3}, ResNet-50 \cite{resnet}, DenseNet-161 \cite{densenet}, MobileNetV2 \cite{mobilenetv2}, EfficientNetV2-medium \cite{efficientnetv2}, ViT-base-16 \cite{vit}, Swin-base \cite{swin}, and ConvNext-base \cite{convnext}. We removed the classification heads, leaving only the feature extractors so that each model outputted a raw feature vector for each image passed in.


\paragraph{Multimodal Models} We selected the pre-trained CLIP ViT-large-14 model \cite{radford2021learning}
as the representative multimodal model for our investigation. We used this model to generate image and text embeddings and logit scores indicating alignment between text-image input pairs. The model implementation was sourced from HuggingFace.

\subsection{Task Paradigms}

\paragraph{Language Model Task}
To estimate the typicality of an exemplar of a category in a language model, we encode the exemplar name as a string, pass it through the language model, and obtain the corresponding word embedding.
We then calculate the cosine similarity, as found by \citeA{bhatia2022transformer} to be the best metric, between this exemplar vector and that of the category prototype, with a higher value indicating that the exemplar is more typical. There are two natural methods for obtaining the prototype of a category: as the average of all exemplar vectors, and as the word embedding obtained by passing the category label to the language model. We explored both methods, finding comparable results. We adopt the former method to maintain consistency with prototype computation for the vision models and with prior work \cite{heyman2019prediction}. 

To evaluate a language model's alignment for a given category, we compute the Spearman correlation between the cosine similarities (representing the typicality judgments of the language model) with the human typicality rankings of exemplars from \citeA{castro21}.

\paragraph{Vision Model Task}


Recall that we collected several images of each exemplar. We pass each image through the vision model, obtain an image embedding, and across these compute the \emph{average} exemplar vector. The prototype vector for the category is defined as the mean of all of these (average) exemplar vectors. We compute the typicality of each exemplar in a category as the cosine similarity between its (average) exemplar vector and the mean prototype vector. We take these as the typicality judgments of the vision model and compute the Spearman correlation with the human typicality rankings from \citeA{castro21}.

\paragraph{Pilot Exploration: Single vs Multiple Image Exemplar Representations}

Choosing to compute the exemplar vector as an \emph{average} of multiple image vectors ensures that our results are not tuned to the choice of a specific exemplar image, and increases the chances that they generalize across images. To justify this approach, we conducted a pilot experiment using the VGG-19 vision model and the Bird category. Using our average vector approach, the Spearman correlation between the model's typicality ratings and those of humans was 0.242. We then examined the consequences of instead using a single image for each exemplar, running 100 trials where we randomly selected one image for each exemplar and recomputed the correlation. These ranged from -0.247 to 0.469, with an average of 0.094. Thus, we conclude that using only one image of each exemplar produces unstable (and artificially low) correlations with human typicality ratings.

\paragraph{Combined Model Task}

To address the third research question, we evaluate each language + vision model pair. Specifically, for each category, we fit a linear model predicting the typicality of an exemplar from its prototype (i.e., the cosine similarity between its vector representation and that of the prototype) in the language model and its prototype in the vision model. We record the (standardized) Beta weight of each predictor variable and evaluate the respective contributions of each modality. We also capture the Spearman correlation coefficient between the predicted rank-ordering of the exemplars and the human typicality ranking from \citeA{castro21}. These values are used, respectively, to assess the respective contributions of language versus vision models in making these predictions and to determine which model pairs perform best in modeling human typicality ratings.

\paragraph{Multimodal Model Task}
The CLIP ViT model produces an embedding for each of the modalities (i.e., vision and text inputs) and outputs a logit score representing their alignment in embedding space. We use text and image-based embeddings to generate category and mean prototypes, mirroring the earlier approaches explored for the language and vision model tasks. We define the category prototype as the embedding of the category input as text and the mean prototype as the average of all representative exemplar image vectors. For the \emph{category prototype approach}, the direct text embedding of the exemplar is taken to be the exemplar representation. For the \emph{mean prototype approach}, the average of all the image embeddings of an exemplar is taken to be the exemplar representation. As with the language and vision tasks, human-model alignment was computed as the Spearman correlation of human typicality ratings and the cosine similarity between exemplar vectors and the corresponding prototype vector for the category (representing model typicality).

We evaluate two additional approaches for this task. The first looks at appending the vectors of different modalities. In this \emph{appended prototype approach}, we define an exemplar representation as the concatenated exemplar representations from the category and mean prototype approaches, producing a vector that is the concatenation of the projections of exemplar text embedding and average exemplar image embedding into joint CLIP embedding space. The typicality alignment is computed as before, with the Spearman correlation between human typicality and the cosine similarity between exemplar and prototype vectors.

In the final, \emph{cross-modality approach} approach, we leverage the CLIP model's computation of logit scores to represent the alignment between image and text representations. 
For this approach, we pass an exemplar image and the category name as an image-text pair input taken by CLIP. The produced logit score represents the alignment between modalities. It is then averaged for all exemplar images to provide the overall alignment score between the exemplar and the category. We calculate the alignment of model and human typicality for each category as the Spearman correlation between the exemplar logit scores and human typicality ratings. 

%
%

\section{Results}

\label{results}

\subsection{Language Models}

\begin{table}[!ht]
    \centering
    \resizebox{0.4\textwidth}{!}{%
    \begin{tabular}{lrr}
    \hline
    Model                   & Mean         & Stdev       \\
    \hline
    all-MiniLM-L12-v2       & \textbf{0.429} & 0.153      \\
    all-mpnet-base-v2       & 0.424        & 0.185       \\
    all-roberta-large-v1    & 0.274        & 0.208       \\
    glove-twitter-200       & 0.421        & 0.186       \\
    sentence-t5-base        & 0.327        & 0.133       \\
    sentence-t5-large       & 0.373        & 0.251       \\
    sentence-t5-xl          & 0.402        & 0.279       \\
    sentence-t5-xxl         & 0.406        & 0.215       \\
    text-embedding-ada-002  & 0.304        & 0.121       \\
    word2vec-google-news-300& 0.222        & 0.166       \\
    xlnet\_base\_cased       & 0.094        & 0.106       \\
    \hline
    \end{tabular}
    }
    \caption{Mean and standard deviation for Spearman correlations across all 27 categories by language model.}
    \label{language_model_statistics}
\end{table}

We first consider the alignment of the language models. Averaging the Spearman correlations across all 27 categories for each model, we observe a range across the models, with a maximum of 0.429 for MiniLM and a minimum of 0.094 for XLNet; see Table \ref{language_model_statistics}. The mean correlation across the models is 0.259 ($SD = .165$). Notably, all models achieve a positive average correlation, signaling general alignment between their predicted typicalities and those of humans.

Although MiniLM performed best among the language models, we note that GloVe, the second oldest model tested, achieved comparable performance ($\rho = 0.421$). This surprising result is consistent with the \cite{bhatia2022transformer} study of the earlier generation language models that we also evaluated: among them, GloVe best accounted for the pairwise (exemplar-exemplar) similarity ratings made by humans. 



\subsection{Vision Models}

\begin{table}[!ht]
    \centering
    \resizebox{0.3\textwidth}{!}{%
    \begin{tabular}{lcc}
    \hline
    Model                   & Mean         & Stdev       \\
    \hline
    alexnet                 & 0.140        & 0.223       \\
    convnext\_base          & 0.058        & 0.186       \\
    densenet161             & 0.051        & 0.165       \\
    efficientnet\_v2\_l     & 0.074        & 0.153       \\
    efficientnet\_v2\_m     & 0.053        & 0.215       \\
    efficientnet\_v2\_s     & 0.081        & 0.207       \\
    inception\_v3           & 0.015        & 0.223       \\
    mobilenet\_v2           & 0.067        & 0.214       \\
    resnet50                & 0.037        & 0.203       \\
    swin\_b                 & 0.058        & 0.226       \\
    vgg19                   & 0.109        & 0.224       \\
    vit\_b\_16              & 0.069        & 0.230       \\
    vit\_h\_14              & \textbf{0.146} & 0.166     \\
    vit\_l\_16              & 0.077        & 0.204       \\
    \hline
    \end{tabular}
    }
    \caption{Mean and standard deviation for Spearman correlations across all categories by vision model.}
    \label{vision_model_stats}
\end{table}

We next consider the alignment of the vision models. Averaging the Spearman correlations across all 27 categories for each model, we see that the vision models show lower alignment than the language models. The average correlation is only 0.0365, ranging from a maximum of 0.1463 (ViT-Huge) to a minimum of 0.0148 (Inceptionv3); see Table \ref{vision_model_stats}.
Like the language models, the vision models all showed positive average correlations, but their small size indicates a much weaker alignment to human typicality ratings.
Paralleling what was found for the language models, there was surprising parity between newer and older models. AlexNet, the oldest vision model architecture we considered, performed nearly as well ($\rho = 0.140$) as the best-performing vision model ViT-Huge ($\rho = 0.146$), and substantially better than VGG-19, the third highest-performing model ($\rho = 0.101$).

\subsection{Combined Models}

\begin{figure}[t]
\centering
\includegraphics[width=1\columnwidth]{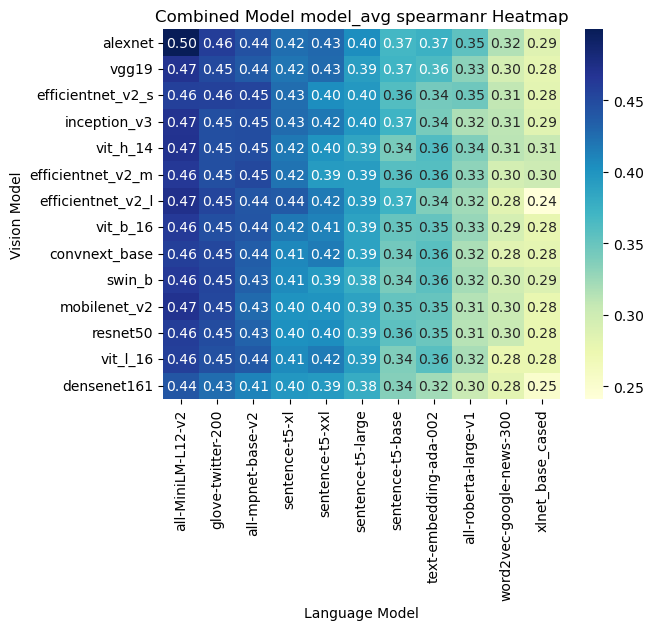}
\caption{For each (language, vision) model combination, the Spearman correlation between its predicted typicalities and the human typicalities, averaged across all categories.}
\label{sorted_heatmap}
\end{figure}

We paired all language models with all vision models and, for each pair, combined the typicality predictions of each model in a linear model to predict the human typicality ratings for each category.
As expected (because adding predictor variables never decreases model fit), the paired models achieved higher correlations than the modality-specific models: compare Figure \ref{sorted_heatmap} against Tables \ref{language_model_statistics} and \ref{vision_model_stats}. Interestingly, the combination of the best-performing vision model (ViT-Huge) and the best-performing language model (MiniLM) did not have the highest correlation with the human typicality data among all of the model pairs. Instead, the combination of MiniLM with AlexNet showed the highest correlation at 0.4995. This indicates that the two models make differential (vs. overlapping) contributions to predicting the typicality of exemplars.

\begin{figure}[t]
\centering
\includegraphics[width=1\columnwidth]{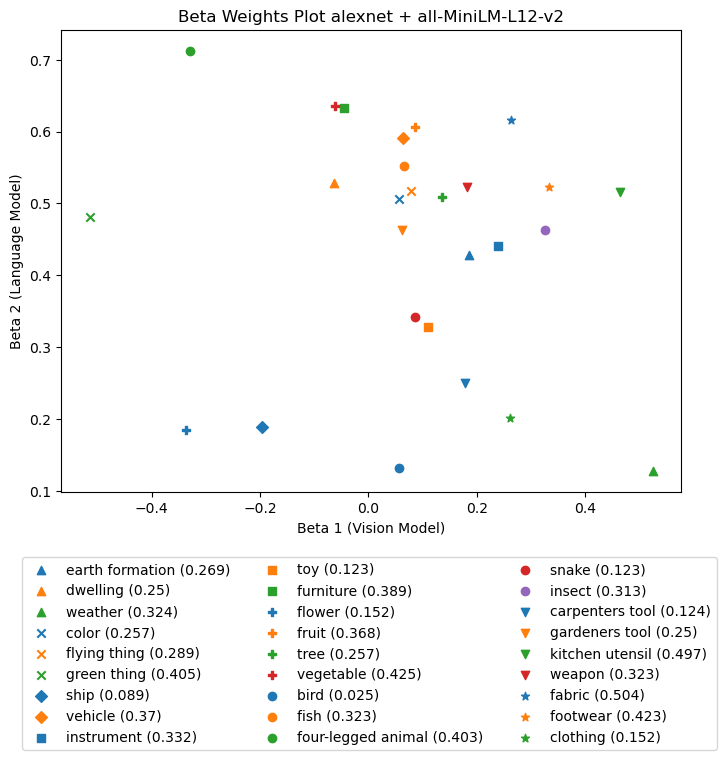}
\caption{Beta weights of the linear models predicting the typicalities of each category for the best-performing combined model (AlexNet + MiniLM).}
\label{top_category_weights}
\end{figure}


Finally, we evaluated whether the typicalities of some categories might be driven more by vision than language. We focused on the best-performing combined model, which is AlexNet + MiniLM. Figure \ref{top_category_weights} shows the Beta weights for each of the linear models predicting the 27 categories. Interestingly, for this combined model, the vision variable contributes to predicting typicality for many of the categories. This is most strikingly the case for \emph{Kitchen Utensil} and \emph{Weather}. 

To aid reader comprehension for Figure \ref{top_category_weights}, we organized the concepts into 7 supercategories and assigned them to a designated shape: Environment (Triangle), Abstract (X), Vehicle (Diamond), Man-Made Miscellaneous (Square), Plant (Plus), Animal (Circle), Man-Made Tool (Upside Down Triangle), and Garment (Star).

\subsection{Multimodal Model}

Table \ref{clip_model_stats} shows the results of the multimodal approaches. The mean prototype approach yields a mean Spearman correlation coefficient that is larger ($r=0.265$) than that of the leading vision model ($r=0.146$), showing that introducing information through alignment with text embeddings was able to align image representations more closely with human concepts.

The performance of the category prototype approach is close to the results of the best language models of similar size, showing that at the very least, introducing information from the vision modality does not adversely affect performance.

The appended prototype approach does not improve alignment with human typicality judgments compared to the category prototype approach. Again, it is worth noting the silver lining here: that explicitly adding exemplar image representations did not harm the overall alignment of exemplar representations with those of humans.

Finally, the cross-modality approach, with its subpar results, suggests that there may exist a fundamentally nonhuman conceptual gap between image and text representations in the joint embedding space of the CLIP model.

\begin{table}[!ht]
    \centering
    \resizebox{0.3\textwidth}{!}{%
    \begin{tabular}{lrr}
    \hline
    Approach & Mean & Stdev \\ \hline
    category           & 0.412       & 0.164              \\
    mean               & 0.265       & 0.169              \\ 
    appended           & \textbf{0.413}       & 0.164              \\ 
    cross-modality       & 0.095       & 0.174              \\ 
    \hline
    \end{tabular}
    }
    \caption{Mean and standard deviation for Spearman correlation across all categories for CLIP ViT.}
    \label{clip_model_stats}
\end{table}


\section{Discussion}

\subsection{Summary of Findings}

The current study addressed four primary research goals. The first was to evaluate whether language models have exemplar representations that show similar typicality gradients as those documented for humans. This was the case, with MiniLM achieving the highest correlation ($\rho = 0.429$) -- one higher than has been observed in prior studies that used earlier-generation language models, older human data sets, and a narrower range of categories  \cite{misra2021language}. The second goal was to ask the same question of vision models. Only one prior study \cite{upadhyay} has addressed this goal, and for only one model (VGG-19) and one category (i.e., Birds). The modest correlation documented there held for the broader range of vision models and categories investigated here. The best-performing model, ViT-Huge, produced typicality predictions that correlated only modestly ($\rho = 0.1463$) with those of humans. The third goal was to examine, for the first time, whether combining language and vision models -- consistent with the multimodal nature of cognition -- produces even better predictions. This was indeed the case, with the AlexNet + MiniLM pairing achieving a 0.4995 correlation with the human typicality ratings.
The final goal was to examine the alignment of an inherently multimodal model with human typicality judgments. We found that there was a sizable correlation between the two, with the mean approach using the vision portion of the model ($\rho = 0.2645$) showing far more promising results than vision models trained on image data alone, and the category approach using the language portion of the model ($\rho = 0.4115$) showing equal if not better results than models trained purely on text data.


Taken together, these findings advance the state of the art in aligning the conceptual representations of ML models and humans. An additional methodological contribution is the creation of a new image set for testing the conceptual alignment of vision models.

\subsection{Limitations and Future Directions}

The limitations of the current study lead naturally to future directions for research, so we discuss both together.

Our investigation of only one multimodal model (CLIP ViT) may not be appropriately representative of the full potential of the multimodal approach, and in more advanced multimodal models, the different modalities may complement each other even more strongly to form better conceptual representations than were observed here. For example, an image of a live chicken should lead to a textual distribution shift where the combined representation of chicken is closer to the representation of a Bird than the representation of chicken as a food in the embedding space. Further work on multimodal contextual alignment for concepts is a goal for future research.

The current study lacks a thorough investigation of large generative language models like GPT-4 \cite{openai2023gpt4} and LLaMA \cite{touvron2023llama} which have shown strong performance for a large variety of tasks.
Future work can investigate prompt-based cognitive modeling of the typicality effects seen in humans using these and similar models.

An important question is why the vision models show lower alignment with human typicality judgments than the language models. A possible explanation for their lower alignment is that a more significant part of image representations are comprised of local information like texture. By contrast, humans rely more on overall shape when making categorization decisions \cite{kurbat2019categorization, rosch1976basic}. Conversely, the superior performance of the language models may be attributed to the text corpora on which they are trained mentioning the attributes of exemplars in the same context as their categories. This may enable them to better capture attribute frequency, which is known to be highly correlated with typicality \cite{rosch75}. Future research should explore these and other explanations for the difference performance levels of language and vision models.

Another limitation may have been our choice to process the images and replace their natural backgrounds with white backgrounds. We did so to avoid the models forming blended representations rather than representations of single exemplars. For example, consider an image of a sparrow (from the Bird category) perched on an oak tree (from the Trees category). However, it is the case that the processed images are different from the naturalistic images on which the models were trained, which may have affected the results of the experiments. In the trade-off between natural images and images with a single object, we chose to focus on the latter. Future research could examine the implications of this choice.

A final limitation concerns the human data on the typicality effect. The \citeA{castro21} study gives typicality rankings for the exemplars (e.g., robin) of categories (e.g., Birds). To derive typicality predictions from the vision models, we sampled ~6-11 images of each exemplar using Google Images and averaged together their vector representations on the final fully-connected layer. To reduce the noise in this average exemplar representation, a future study could collect human typicality ratings on the exact images provided to the vision models. This would enable less noisy evaluation of the typicality gradients of vision models and potentially increase their alignment to human typicality rankings.

\bibliographystyle{apacite}

\setlength{\bibleftmargin}{.125in}
\setlength{\bibindent}{-\bibleftmargin}

\bibliography{cogsci}

\end{document}